\documentclass[%
 reprint,
superscriptaddress,
 amsmath,amssymb,
 aps,
]{revtex4-1}
\usepackage{color}
\usepackage{amssymb}
\usepackage{amsmath}
\usepackage{bm}
\usepackage{graphicx}
\usepackage{epstopdf}
\usepackage{newtxtext,newtxmath}
\usepackage{xfrac}
\usepackage{amsfonts}
\usepackage{xspace}
\usepackage{multirow}
\usepackage{dcolumn}% Align table columns on decimal point

\begin{document}

%\preprint{APS/123-QED}

\title{Improving Demand Forecasting in Open Systems with Cartogram-Enhanced Deep Learning}% Force line breaks with \\
%\thanks{A footnote to the article title}%

\author{Sangjoon Park}
% \email{sang.joon.park.96@gmail.com}
 \affiliation{%
 Department of Applied Artificial Intelligence, Hanyang University, Ansan 15588, Korea
}%
\author{Yongsung Kwon}
 \affiliation{%
 Department of Applied Artificial Intelligence, Hanyang University, Ansan 15588, Korea
}%
\author{Hyungjoon Soh}
\affiliation{%
 Department of Applied Physics, Hanyang University, Ansan 15588, Korea
}%
 \affiliation{%
  Department of Physics Education, Seoul National University, Seoul 08826, Korea
}%
\author{Mi Jin Lee}
\email{mijinlee@hanyang.ac.kr}
 \affiliation{%
 Department of Applied Physics, Hanyang University, Ansan 15588, Korea
}%
\author{Seung-Woo Son}%
 \email{sonswoo@hanyang.ac.kr}
  \affiliation{%
 Department of Applied Artificial Intelligence, Hanyang University, Ansan 15588, Korea
}%
 \affiliation{%
 Department of Applied Physics, Hanyang University, Ansan 15588, Korea
}%

\date{\today}% It is always \today, today,
             %  but any date may be explicitly specified

\begin{abstract} 
Predicting temporal patterns across various domains poses significant challenges due to their nuanced and often nonlinear trajectories. To address this challenge, prediction frameworks have been continuously refined, employing data-driven statistical methods, mathematical models, and machine learning. Recently, as one of the challenging systems, shared transport systems such as public bicycles have gained prominence due to urban constraints and environmental concerns. Predicting rental and return patterns at bicycle stations remains a formidable task due to the system's openness and imbalanced usage patterns across stations. In this study, we propose a deep learning framework to predict rental and return patterns by leveraging cartogram approaches. The cartogram approach facilitates the prediction of demand for newly installed stations with no training data as well as long-period prediction, which has not been achieved before. We apply this method to public bicycle rental-and-return data in Seoul, South Korea, employing a spatial-temporal convolutional graph attention network. Our improved architecture incorporates batch attention and modified node feature updates for better prediction accuracy across different time scales. We demonstrate the effectiveness of our framework in predicting temporal patterns and its potential applications.

%\begin{description}
%\item[PACS numbers]
%05.45.Tp, 87.23.Kg, 89.20.Bb
% 05.45.Tp: Time series analysis
% 87.23.Kg: Dynamics of evolution
% 89.20.Bb: Industrial and technological research and development
% PACS numbers 2020 (https://ufn.ru/en/pacs/)
%\item[Keyworks] Cartogram, Graph Neural Network, Demand Forecasting
%\end{description}
\end{abstract}

%\pacs{Valid PACS appear here}% PACS, the Physics and Astronomy
                             % Classification Scheme.
%\keywords{Suggested keywords}%Use showkeys class option if keyword
                              %display desired
\maketitle

%\tableofcontents

\section{Introduction}
\label{sec:introduction}
Predicting temporal patterns of a system has been one of the most challenging tasks in diverse fields and research topics, such as financial crisis~\cite{BERG1999561,BUSSIERE2006953, doi:10.1061/JCEMD4.COENG-13959,doi:10.1089/big.2020.0158}, outbreaks of the recent pandemic~\cite{FANELLI2020109761, 10.1371/journal.pcbi.1008431, PhysRevX.10.041055}, and cultural or industrial popularity~\cite{10.1145/1772690.1772754, 10.1145/2802558.2814645, LYMPEROPOULOS2016585, PhysRevE.91.012815}, which is because of their nuanced and often nonlinear trajectories. Despite the inherent complexity, the paths to building forecast frameworks of predicting the future patterns to make informative decisions, optimize processes, and mitigate risks have been steadily polished and developed in the realm of each field, such as data-driven statistics methods~\cite{10.1111/jifm.12053,doi.org/10.1002/csr.2567, Shashank_2021} and mathematical models with the mean field approach~\cite{FANELLI2020109761, PhysRevX.10.041055, LYMPEROPOULOS2016585, PhysRevLett.92.178701, PhysRevE.99.032309}. % due to its essentiality and importance.
%These tasks also occur in many situations in our daily lives.

Shared transportation systems such as public bicycle and car sharing recently have become one of the growing systems% from various points of view
. For a shortage of land for parking lots in populated cities as well as environmental conservation such as carbon neutrality, using the micro-mobility and car sharing has been getting popular. The convenience of letting individual users freely control the rental and return at any station, differently from traditional transportations, also promotes the growth of shared transportation systems. Therefore, it is crucial to predict the rental and return patterns at the stations for stable operation. 

However, predicting the temporal patterns of the rental and return is still challenging for the following reasons: (i) It is an open system---the total number of users fluctuates, and the installation and shutdown of the stations are more frequent than the conventional systems. (ii) There exists an imbalance of rental and return between stations. Due to such uncertainty, some machine learning models have been designed for the prediction of ``rental'' or ``return'' (collectively called ``{demand}'' in this study) at a station  level~~\cite{Lin2018shared_transportation1, Yao2018ai_example1, Zhang2020ai_example2}. Specifically, to utilize temporal and spatial patterns, previous studies have adopted either the recurrent neural network or convolution neural network (CNN)~\cite{Zhai2021cnn_example1, Feng2020cnn_example2} or both simultaneously~\cite{Wang2021cnnrnn_example1,Wang2020cnnrnn_example2}. However, the CNN is limited to capturing information on adjacent regions only, naturally leading to the dismissal of long-range correlations between geographically distant regions with similar characteristics such as floating population and facility density. This limitation can be solved by introducing a graph neural network (GNN) that uses graph information between regions, accompanying the demanding computation~\cite{Kipf2016gnn_example1, Lee2019gnn_example2}. To overcome computational complexity, many studies have adopted coarse-grained demand by aggregating some demands at stations for a given window~\cite{Zhang2017regional_demand_example1,Li2019regional_demand_example2,Kim2019regional_demand_example4}. Yet, the coarse-graining process could blur the regional characteristics (for instance, the distribution and pattern of demand).

In this study, to secure both the lower complexity and higher accuracy, we introduce the cartogram approaches obtained by iterating the Voronoi tessellation~\cite{VoronoiTessellation}. The cartogram is a distorted map based on a feature of interest. The map of stations' locations is distorted by spreading the stations' positions using the Voronoi tessellation, until the station density of each Voronoi cell, equivalent to the inverse size of the Voronoi polygon, becomes homogeneous. As a result, stations of similar characteristics get clustered nearby, which is evidenced by the correlation coefficient. Furthermore, the uniform spatial distribution enables us to predict the new demand for newly installed stations that did not appear in the training data, which has not been accomplished so far. The absence-intraining-data but presence-in-test-data has hindered the long-time scale
prediction. However, the successful prediction of brand-new demand naturally enables us to overcome such short-term predictions. The mean-field-like approach is in line with the fact that machine learning and physics have complementarily brought the advancement of each field~\cite{carleo2019machine}.

We explore the demand (rental and return) data of public bicycles in Seoul (the capital city of South Korea)~\cite{station_usage_dataset}: year 2018 for training data and 2019 for test data (prediction). To utilize and predict the spatio-temporal patterns, we employ a spatial-temporal convolutional graph attention (ST-CGA) network ~\cite{Zhang2020STCGANetwork} that consists of mainly three parts as self-attention~\cite{Vaswani2017selfattention}, graph attention network~\cite{Velickovic2017GAT}, and CNN. For efficiency and improved performance, we consider three different time scales of an hour, a day, and a week, and modify the self-attention into the batch attention and the node-feature update in the graph attention network, compared with the ordinary ST-CGA network. In particular, batch attention refers to various data at different times, which leads to contemplating the temporal correlation, and then we accomplish multiple prediction results at once with higher accuracy. We believe that our framework is applicable for predicting temporal patterns even for untrained spatial data.
%, and it could be ubiquitously applied with better performance.

The rest of this paper is organized as follows: In Sec.~\ref{sec:data_method}, we introduce the empirical data and the ST-CGA model, utilizing the cartogram approaches. Especially, we focus on the modification in the ST-CGA model and the effects before and after applying it to the cartogram idea. In Sec.~\ref{sec:results}, we showcase the prediction performance overall and the initial demand prediction of a newly installed station. Lastly, we summarize this paper and provide a discussion in Sec.~\ref{sec:conclusion}.

\section{Data construction and Prediction method}
\label{sec:data_method}
To understand spatio-temporal patterns such as regional demand forecasting, many studies have used deep learning models such as a combination of the recurrent neural network (processing well in time series) and CNN (doing well in images)~\cite{Zhai2021cnn_example1,Wang2021cnnrnn_example1,Lee2019gnn_example2}.

In this study, we analyze and predict the spatio-temporal demand patterns of public bicycles in Seoul, Korea, which is definitely an open system. The rental or return patterns are spatially heterogeneous across the rental stations, as seen in Figs.~\ref{fig:mean_x}(a) and~\ref{fig:mean_x}(b), so we also use CNN for prediction. However, the CNN only factors in the adjacent regions inherently limited by the size of its spatial window, which is called a filter, although the return-and-rental sometimes manifests itself between distant stations. Furthermore, some regions have similar characteristics, such as population density and land use, which could affect the demand pattern, even if the rental-and-return among them does not happen indeed. To contemplate such a long-range correlation, the graph neural network and relevant variants using network information have been developed. Among others, we exploit the ST-CGA network~\cite{Zhang2020STCGANetwork} that embraces both characteristics of the convolutional and graph neural networks, which we also modify to enhance precision. In this section, we describe the conversion of the empirical data into a suitable form as input data and an entire architecture of the prediction model, focusing on our modification.

\subsection{Rental-and-return data in Seoul}
\label{subsec:data}
%%% 지역수요 생성
%%M = 17, N = 15
%% 2018년 대여소 수 : 1538개
%% 2019년 대여소 수 : 1554개
%% 2018년 1월 1일 ~ 2019년 12월 31일
% Fig: time resolution에 따른 수요 패턴
\begin{figure}[t]
\centering
\includegraphics[width=\columnwidth]{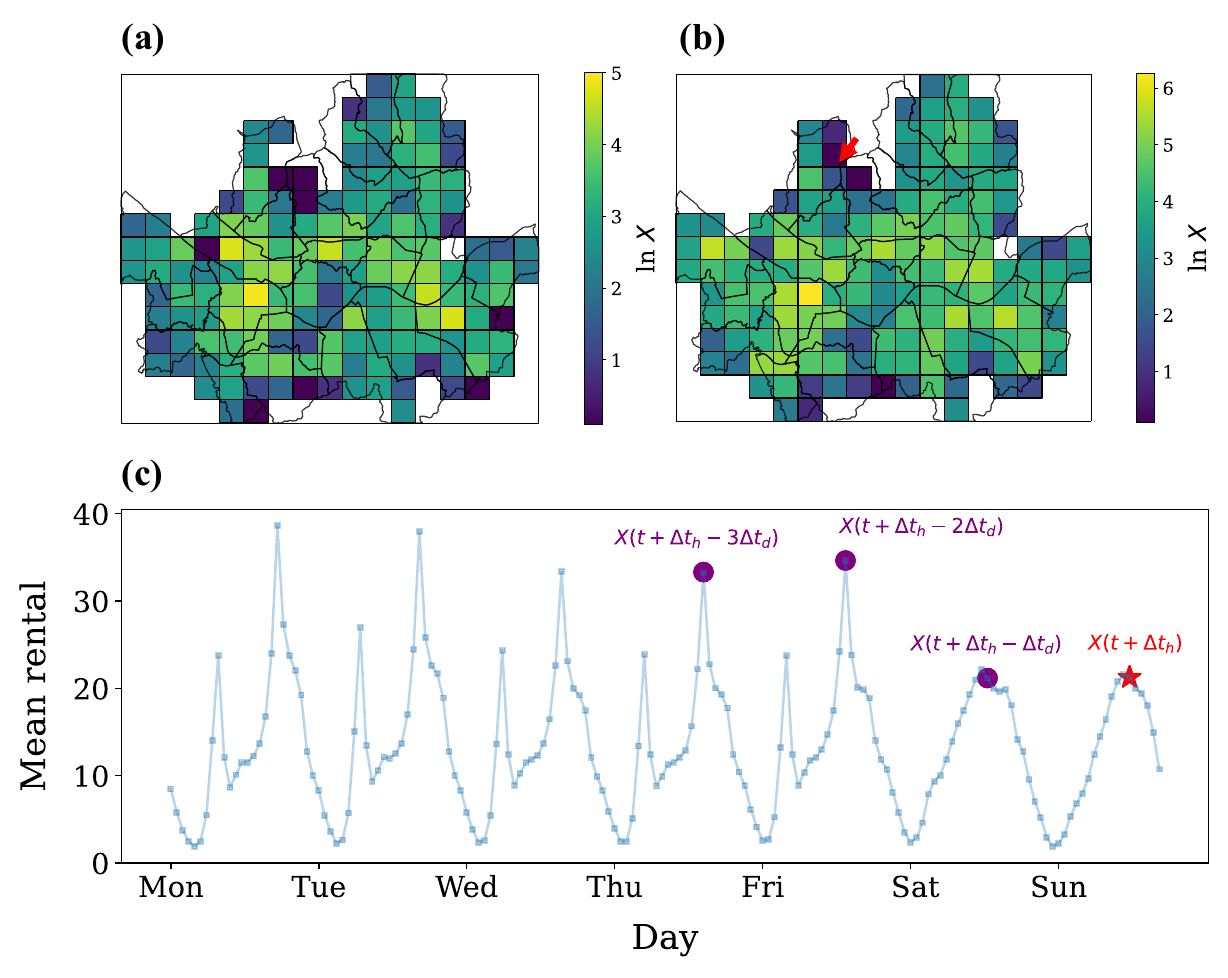}
\caption{The usage of public bicycles in Seoul. Snapshots of rentals (one of the two types of demand) (a) at 18:00, June 3rd, 2018 and (b) at 18:00, June 3rd, 2019. The sum $X$ of rentals in a cell on a grid is indicated by color on a logarithmic scale. Note that a cell marked by a red arrow solely has a newly installed station in year 2019, of which the administrative-gu is Eunpyeong-gu. (c) A time series of rentals in the year 2019. The mean value spanning a week is plotted as guidance to showcase the quasi-periodicity (we actually use the original sequence, not this mean value). As an illustrative example, we mark the $\tau_d$ points of rentals $X$ required to predict $X(t+\Delta t_h)$, when considering a day resolution $\Delta t_d$.
}
\label{fig:mean_x}
\end{figure}

We collect time series data for rentals and returns every hour from 0:00-0:59 on January 1, 2018, to 23:00-23:59 on December 31, 2019~\cite{station_usage_dataset}. The data is recorded at a minute-level unit, which exhibits noisy patterns due to factors such as human errors (record omissions) and drastic weather changes. To mitigate this noise, we aggregate the data from minute-level to hourly intervals. Thus, we have the time stamp [(365 days)$\times$(24 hours)] for two years, so $t\in \mathbb{T}$ with $\mathbb{T}=\{1, 2, \cdots, 8760 \}$ for a given year (that is, $t=1$ stands for 0:00 on January 1, and $8760$ for 23:00 on December 31 of the same year). We will use the empirical data for the year 2018 as a training data set and those for the year 2019 as a test data set. 

Let us refer to the rental or return as \textit{demand} collectively. The total number of rental stations is 1,538 for 2018 and 1,554 for 2019. The map image containing information on rental or return is necessary for CNN. To reduce computational complexity, we divide the city map into a $M\times N$ grid with $M=17$ and $N=15$ by a $2~\mathrm{km}$ resolution, giving the $M\times N=255$ cells, and then use the coarse-grained demand for a cell $i$ at time $t$ as
\begin{equation}
    X_{i}(t) = \sum_{j \in \mathbb{R}_{i}} x_{j}(t),
\label{eq:xi}
\end{equation}
where the raw data $x_j(t)$ is the demand (rental or return) of a station $j$ located at a cell $i$ ($i=1, 2, \cdots, 255$), and $\mathbb{R}_i$ is a set of the stations within a cell $i$ [see Figs.~\ref{fig:mean_x}(a) and~\ref{fig:mean_x}(b)]. 

%%% time resolution에 따른 수요 패턴 정의
To understand the characteristics of the temporal patterns, we display the averaged pattern of hourly rental, over cells and for the year 2019 in Fig.~\ref{fig:mean_x}(c). %We do not show the mean yearly pattern for showing the example of $\bar{X}^{w}$ because it is too lengthy to be plotted here.
The quasi-periodic behaviors in different time scales are shown, e.g., rush hour/non-rush hour, weekday/weekend, and seasonal effect (although not shown here but straightforward to be expected), which is persuasive for taking the various time scales to seize this quasi-periodicity for training.  

% Fig: 모델 구조
\begin{figure*}[t]
\includegraphics[width=1.0\textwidth]{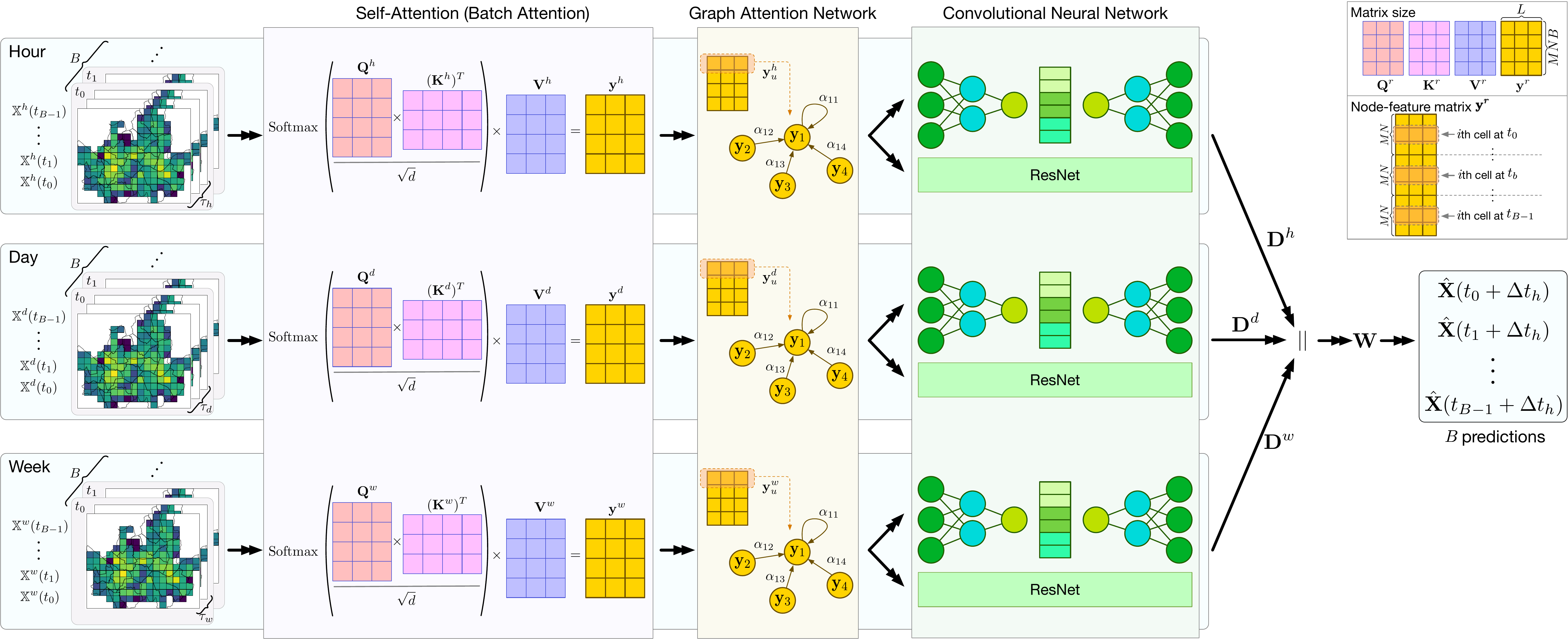}
\caption{The architecture of the training process by the ST-CGA network model. We construct the demand dataset $\mathbb{X}(t_b)$ which contains the demand for $M\times N$ cells at $B$ different pivot time $t_b$'s with various temporal resolutions in Eq.~(\ref{eq:input_x}). The self-attention (in fact, batch attention) process (see Appendix~\ref{seca:selfattention} for notation) offers the node feature matrix between cells (nodes). The all-to-all structure in terms of the network becomes an input of the graph attention network, and graph attention first provides attention scores $\alpha$ regarded as the link weights of the pair of nodes. The node feature $\mathbf{y}_u$ is updated by the attention scores and learnable important matrix.  As an example, herein we illustrate the ego-centric connection structure. The input data with each temporal resolution are trained in parallel and then merged after the CNN. After some iterations with backpropagation, we finally obtain the prediction values at $t_b+\Delta t_h$, i.e., $\hat{\mathbf{X}}(t_b+\Delta t_h)$.}
\label{fig:model}
\end{figure*}
%%% 원래는 all-to-all이 맞고, 여기에선 1번 노드에 대해서 ego-centric한 것만 남김 

Utilizing the coarse-grained demand in Eq.~(\ref{eq:xi}), we construct the temporal sequence for various time resolutions $\Delta t$ such as an hour, a day, and a week. Temporal resolutions are denoted by $\Delta t_h\equiv1\mathrm{h}$ for one hour, $\Delta t_d\equiv1\mathrm{d}$ for one day, and $\Delta t_w=1\mathrm{w}$ for one week, so $\Delta t_d =24\Delta t_h$ and $\Delta t_w =7\Delta t_d$ for consistency. In this study, we predict the demand after one hour from the pivot time $t$, that is, $X_i(t+\Delta t_h)$. The temporal sequence for training for a given resolution $\Delta t_r$ with $r\in \{h, d, w\}$ is built as
\begin{equation}
    \mathbf{X}_i^{r}(t;\tau_r)=[X_i(t+\Delta t_h-\Delta t_r), \cdots, X_i(t+\Delta t_h-\tau_r \Delta t_r)]^{T},
\label{eq:x_subset}
\end{equation}
where $\tau_r$ is a hyperparameter for the truncation, and the superscript $T$ means the matrix transpose. We heuristically select $\tau_h=3$, $\tau_d=3$, and $\tau_w=2$. Note that the last demand with an hour resolution is nothing more than $X_i(t+\Delta t_h-\Delta t_h)=X_i(t)$. We compose a set by aggregating the sequences in Eq.~(\ref{eq:x_subset}) of all $M\times N$ cells for a given temporal resolution $r$ as
\begin{equation}
    \mathbb{X}^{r}(t)=\{\mathbf{X}_1^{r}(t; \tau_r), \cdots, \mathbf{X}_i^{r}(t; \tau_r), \cdots, \mathbf{X}_{MN}^{r}(t; \tau_r)\},
    \label{eq:input_x}
\end{equation}
and always consider the triplet $[ \mathbb{X}^h(t), ~\mathbb{X}^d(t),~ \mathbb{X}^w(t)]$ as \emph{input} data unit in our machine learning model. For clarity, in this paper, the typefaces $A$, $\mathbf{A}$, and $\mathbb{A}$ stand for a scalar value, a matrix (and column vector), and a set, respectively.

% % Fig: Voronoi tessellation 예시
% % cumulative (2019년 중간에 사라진 대여소들도 모두 포함)
% \begin{figure}[t]
% \centering
% \includegraphics[width=0.9\columnwidth]{imgs/tessellation example.pdf}
% \includegraphics[width=0.9\columnwidth]{imgs/cartogram result.pdf}
% \caption[In 2019 Seoul public bike stations, using the cartogram method before and after.]{In 2019 Seoul public bike stations, using the cartogram method before and after. (a) The stations in the center of Seoul are more dense than those in the outskirts. (b) After using the cartogram method, the stations are more uniform in space.}
% \label{fig:tessellation example}
% \end{figure}

\subsection{Spatio-temporal convolutional graph attention network}
\label{subsec:model}

We illustrate an architecture of the ST-CGA network in Fig.~\ref{fig:model}. The ST-CGA network has three compartments as the self-attention~\cite{Vaswani2017selfattention}, graph attention network~\cite{Velickovic2017GAT}, and CNN. We modify the model structure to improve the prediction for $\hat{X}_i(t+\Delta t_h)$ (the hat notation indicates a predicted value). The set $\mathbb{X}^{r}(t)$ of time sequences in Eq.~(\ref{eq:input_x}) for every resolution $r$ experiences the learning process via self-attention, graph attention, and CNN in parallel, and then $\mathbb{X}^{h}(t)$, $\mathbb{X}^{d}(t)$, and $\mathbb{X}^{w}(t)$ are merged after completing CNN. The detail of the ST-CGA network is well described in Ref.~\cite{Zhang2020STCGANetwork}, so we briefly furnish the revised parts in the first two compartments as the focal points and then describe how to measure the performance. 

\emph{Self-attention} or \emph{batch attention}: A set $\mathbb{X}^{r}(t)$ of demand vectors is used as input and all possible pairwise correlation are evaluated in a latent space to attain the predicted demand value $\hat{\mathbf{X}}(t+\Delta t_h)\equiv[\hat{X}_1(t+\Delta t_h), \hat{X}_2(t+\Delta t_h), \cdots, \hat{X}_{MN}(t+\Delta t_h)]^{T}$. It is called the self-attention; only referring to the current data itself, relevant to the pivot time $t$ in this study (i.e., $\mathbb{X}^{r}(t)$ tautologically). In self-attention, the cell-to-cell (spatial) correlation and the temporal correlation within the time interval $\tau_r$ are concerned.  The previous study~\cite{Cheng2021batch_attention} has revealed that the extended version of self-attention improves prediction performance. Referring to different data together is called {\it batch attention}. In this paper, the ``different data'' indicate the data at the different pivot time $t^{\prime}$ ($t\neq t^{\prime}$). In other words, adopting batch attention means considering the mixture of the cell-to-cell and temporal correlation in a wide range of time stamps represented by different pivot times.

The number of the data to be referred corresponds to a batch size $B$, and we set this hyperparameter $B=16$. We randomly choose the $B$ different pivot times, and a bundle of $\mathbb{X}^{r}(t_b)$'s (also $t_b\in\mathbb{T}$ and $b=0, 1, \cdots, B-1$) becomes a new input unit. When $B=1$, it returns to the self-attention. Utilizing batch attention enables us to attain the $B$ predicted values $\hat{\mathbf{X}}(t_b+\Delta t_h)$ at once at the final stage. We adopt the batch attention but still use the terminology {\it self-attention} by convention.

This self-attention yields a $MNB\times L$ matrix $\mathbf{y}^{r}$ with $L$ being dimensions of the latent space and a hyperparameter (see Appendix~\ref{seca:selfattention}). The vector of the $u$-th row of the matrix is denoted as $\mathbf{y}^{r}_u$, where $u=i+b\times MN$, stores $L$ features of a cell $i$ at a pivot time $t_b$, which embodies all the influences of the other $v$-th rows, where $v= i^{\prime}+b^{\prime}\times MN$, calculated in the latent space. Although we cannot interpret the exact physical meaning of the features as always in deep learning, this matrix is often called the correlation matrix since it carries some relatedness among cells across times. The detail is described in Appendix~\ref{seca:selfattention}.

\emph{Graph attention network}: The graph attention network requires a graph structure represented by an adjacency matrix, but our empirical data do not provide the spatio-temporal adjacency matrix. Therefore, we assume the globally coupled connection between cells (coarse-grained nodes) and then construct a weighted adjacency matrix $\mathbf{A}^r$ from the correlation matrix $\mathbf{y}^{r}$ by the attention mechanism~\cite{Velickovic2017GAT}. An element of the asymmetric weighted adjacency matrix, also known as an attention score, say $\alpha_{uv}$, is evaluated by 
\begin{equation}
    \alpha^{r}_{uv} = \frac{\exp\left(\mathrm{LeakyReLU}(\mathbf{W}_0^r[\mathbf{W}_0^r\mathbf{y}^{r}_u||\mathbf{W}_0^r\mathbf{y}^{r}_v])\right)}{\sum_{k}\exp\left(\mathrm{LeakyReLU}(\mathbf{W}_0^r[\mathbf{W}_0^r\mathbf{y}^{r}_u||\mathbf{W}_0^r\mathbf{y}^{r}_k])\right)},
\label{eq:attention}
\end{equation}
where $\mathbf{W}_0^r$ is an importance matrix and learnable through learning process, $\mathrm{LeakyReLU}(x)=\max[(\mathrm{constant}<1)*x, x]$ is a popularly-used activation function, and $||$ denotes the concatenation operator [e.g., for two matrices 
$\mathbf{a}=\big(\begin{smallmatrix}
  a_1 \\
  a_2 
\end{smallmatrix}\big)$ and 
$\mathbf{b}=\big(\begin{smallmatrix}
  b_1 \\
  b_2 
\end{smallmatrix}\big)$, then
$\mathbf{a}||\mathbf{b}=(\begin{smallmatrix}
  a_1 & a_2 & b_1 & b_2 \end{smallmatrix})^{T}$].

%%% 변형한 GAT 업데이트 수식, 사용한 수식과 유사한 선행연구 (APPNP)
There are three layers in the graph attention network (not shown in this paper because the detailed description of its basic structure is not our main concern), and the attention score is acquired only at the first layer. Whenever the layer passes, every node feature is updated as $\mathbf{y}^{r \prime}_u = \sum_v \alpha^{r}_{uv}\mathbf{y}^{r}_u\mathbf{W_0}$ across all three layers of the graph attention network in general.  In our modification, we update the node feature without the importance matrix $\mathbf{W_0}$, except for the first layer, as 
\begin{equation}
    \mathbf{y}^{r \prime}_{u} = \sum_{v}\alpha^{r}_{uv}\mathbf{y}^{r}_{u}.
\label{eq:updated_nodefeature}
\end{equation}
This modified update process has been verified in the previous study on an approximate personalized propagation of neural predictions~\cite{Gasteiger2018APPNP}. We conjecture that the structure of the graph attention network considered here may be similar to that of the approximate personalized propagation, supported by the improved performance.

%%% 모델의 마지막 부분과 손실함수
\emph{Predicted value}: The updated node features $\mathbf{y}_i^{\prime}$'s lead to the final output matrices $\mathbf{D}^h$, $\mathbf{D}^d$, or $\mathbf{D}^w$ of the CNN, whose important channels are determined by the feed-forward neural network. Merging the three matrices, we accomplish a final output $\hat{\mathbf{X}}$ using another learnable importance matrix $\mathbf{W}$ computed as 
\begin{equation}
\hat{\mathbf{X}} = \left(\mathbf{D}^{h}||\mathbf{D}^{d}||\mathbf{D}^{w}\right)\mathbf{W}.
\label{eq:final_output}
\end{equation}
Measuring the mean squared loss in the learning process is computed as 
\begin{equation}
L = \sum_{b=0}^{B-1} \sum_{i=1}^{MN} \left[\hat{X}_i(t_b+\Delta t_h)-X_i(t_b+\Delta t_h)\right]^2,
\label{eq:loss}
\end{equation}
and iterating all the process from the self-attention by back-propagation, then we finally obtain the predicted demand (rental or return) $\hat{X}_i$.

\subsection{Cartogram approaches to spread the stations}
\label{subsec:cartogram}

\begin{figure}
\centering
\includegraphics[width=\columnwidth]{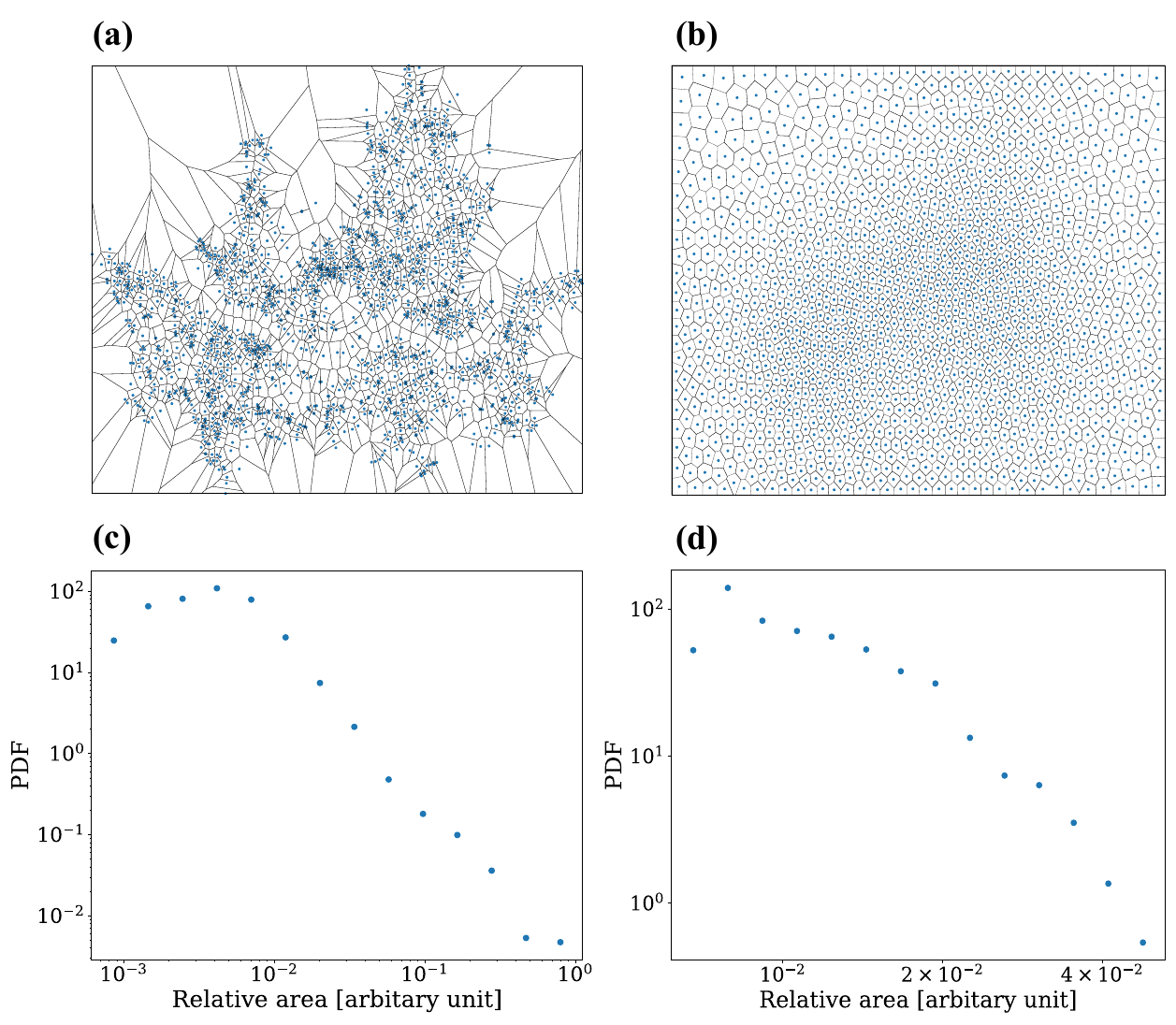}
\caption{The original map and cartogram with the Voronoi tessellation. The station within a Voronoi cell and the relative area distribution of the original empirical data [(a), (c)] and those of the cartogram [(b), (d)]. (a), (b) every polygon (Voronoi cell) contains only one station. (c), (d) The distribution of the polygon's relative area. The relative area is obtained by scaling the largest area in (a).}
\label{fig:cartogram}
\end{figure}

%%% 우리가 카토그램을 사용하는 이유
We pay attention to an open system confronted with a varying size of the system. The rental-and-return system also undergoes variations in demand either by the number of users or by the installation or shutdown of the stations. The absence of demand in a cell means null input data for training, which hinders us from predicting its future demand. To overcome this problem, we uniformly spread the stations across the map like a cartogram, so that empty cells do not exist. The cartogram is a thematic map that contains demographic or other features and yields the visually distorted geographic size, proportional to the relative level of feature of interest~\cite{Sun2010cartogram_example, cartogram_example2}. In the sense that we visually distort a map suitably to our purpose -- homogeneous distribution of the stations, the result of the following method using a tessellation is a cartogram based on the facility density. 

%%% 카토그램 과정 설명
We build up the process to spread the stations as follows, using a Voronoi cell that is a polygon containing only one point~\cite{Aurenhammer1991cartogram} as illustrated in Fig.~\ref{fig:cartogram}: (i) Make the triangles fully packed in the plane by connecting the three nearest points (stations), without any cross lines and overlap among the triangles. (ii) For every triangle, form a circumcircle and then find the center of the circle. 
% (iii) Form new polygons by connecting the centers of the circles without any cross lines and overlap among others, and position the points at the centers of the polygons that they belong to. We iterate the steps (i)-(iii) until there is no significant change in the spatial distribution of the stations, then any empty cell is absent in the new map divided into the $M\times N$ grid.
(iii) Form new polygons by connecting the centers of the circles without any cross lines and overlap among others and then find the center of the polygon. (iv) Move the positions of the points to the centers of the newly formed polygons that they belong to. We iterate the steps (i)-(iv) until there is no significant change in the spatial distribution of the stations, then any empty cell is absent in the new map divided into the $M\times N$ grid.
The new cells after sufficient iterations compose the input data $\mathbb{X}$ in Eq.~(\ref{eq:input_x}). It can be seen that the areal distribution of the polygons becomes homogeneous after iteration in Figs.~\ref{fig:cartogram}(c) and~\ref{fig:cartogram}(d). The homogeneity allows us to predict the demand of a newly installed station, in a similar spirit of a mean-field approach.

% Fig: 카토그램 사용 전과 후, 각 지역의 변동계수와 피어슨상관계수 측정
\begin{figure}
\includegraphics[width=1.0\columnwidth]{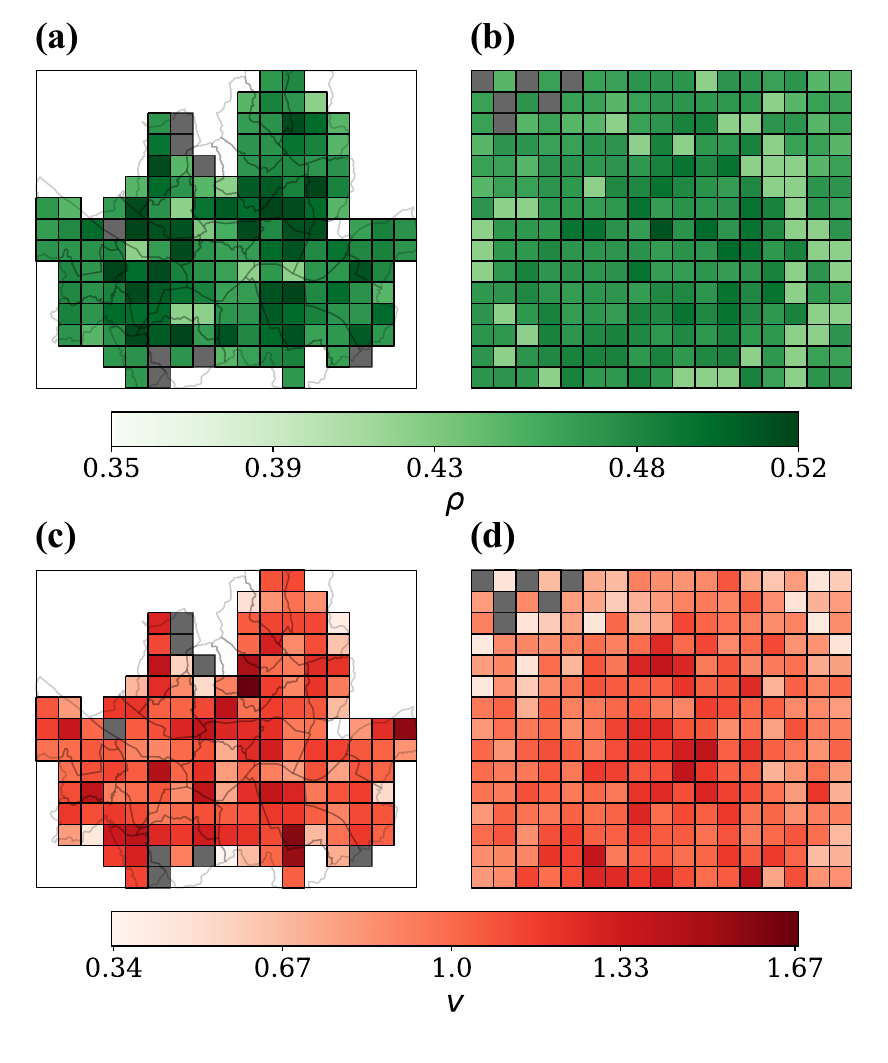}
\caption{The similarity of the demand patterns of stations within a cell before [(a), (c)] and after [(b), (d)] using the cartogram (representatively, the rental for the year 2019). (a), (b) Pearson's correlation coefficient $\rho$ for every cell. (c), (d) The coefficient of variation $v$. The cells with $\rho=1$ and $\nu=0$ colored in gray have only one station each.}
\label{fig:CV_P}
\end{figure}

%%% coefficient of variation 측정 결과
We use the resultant cartogram to infer a demand of a new station with the help of the adjacent demands, under the assumption that the characteristic of the demand is similar to that of the adjacent stations. To verify such a small dispersion of demands in every cell in the cartogram map, we measure the Pearson correlation coefficient after detrending and the coefficient of variation of a cell $i$ for a given year. The correlation value $\rho$ captures the temporal similarity between pairwise stations within a cell. The correlation in the cartogram map is still as positive as in the original map, giving $0.52\pm 0.1$ and $0.49\pm 0.08$, respectively [Figs.~\ref{fig:CV_P}(a) and ~\ref{fig:CV_P}(b)]. Our suggested iteration method induces the gradual spreading of stations that may have similar characteristics of demands, in careful consideration of spatial adjacency, keeping the correlation level.

The dispersion of the demands among the stations in a cell is measured by the coefficient of variation as 
\begin{equation}
v_{i}=\frac{1}{T}\sum_{t\in \mathbb{T}} \frac{\sigma_{i}(t)}{\mu_i(t)}, 
\label{eq:cv}
\end{equation}
with the mean $\mu_i(t)=\frac{1}{|\mathbb{R}_i|}\sum_{j\in \mathbb{R}_i} x_j(t)=\frac{1}{|\mathbb{R}_i|}X_i$ and standard deviation $\sigma_i^2(t)={\frac{1}{|\mathbb{R}_i|}\sum_{j\in \mathbb{R}_i} \left[x_j(t)-\mu(t)\right]^2}$ at time $t$ ($|\cdots|$ is a cardinality of a set). The coefficient of variance in the original map and the cartogram map for the year 2019 as an example is displayed in Figs.~\ref{fig:CV_P}(c) and~\ref{fig:CV_P}(d). The mean $v$ is $1.0\pm 0.3$ for the original map, while $0.9\pm 0.2$ for the cartogram. Thus, our uniformization keeps the correlation and the dispersion of the demands similar to the original level, although the number of stations in cells and the number of cells are varied. We also remark that empty cells disappear on the map as seen in Figs.~\ref{fig:CV_P}(b) and~\ref{fig:CV_P}(d), which potentiates the demand prediction of the new station solely installed in the location marked by an arrow in Fig.~\ref{fig:mean_x}(b). This significant increase in the input data point $X_i$, without violating the statistical level, is expected to enhance the accuracy of predictions.

\section{Results}
\label{sec:results}
We train the empirical rental and return data separately for the year 2018 using the model framework in Sec.~\ref{sec:data_method}, and then predict the demands for the year 2019 and compare the predicted values with the empirical data. We first analyze the meaning of the attention score in Eq.~(\ref{eq:attention}), then verify the performance of this prediction framework and argue the demands of a station that did not exist in 2018 but was newly installed in 2019. 

\subsection{Analysis of the attention score of the graph attention network}
\label{subsec:attention_score}
% Fig: GAT의 어텐션 스코어 클러스터링 & 각 클러스터에 속하는 대여소의 원래 위치
\begin{figure*}
\includegraphics[width=1.0\textwidth]{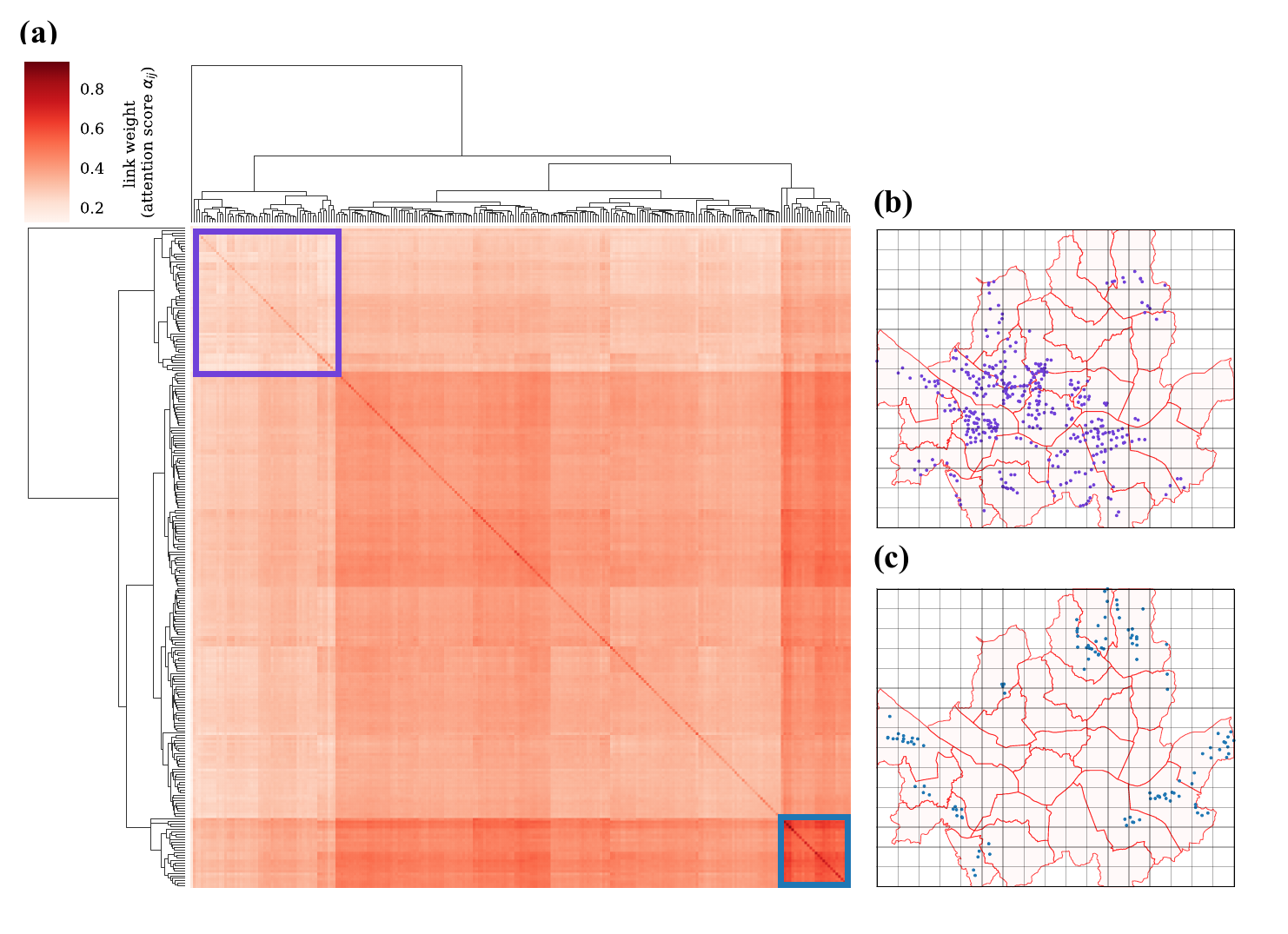}
\caption{The analysis for the attention score between every cell pair in the graph attention network, for the training data (year 2018). (a) The yearly average attention score, or weighted adjacency matrix. The 255 cells correspond to a (coarse-grained) node, and the average attention score $\alpha_{ij}$ represents the weight of the link. The higher $\alpha_{ij}$ has a darker color. Clusters with low and high $\alpha$'s are surrounded by the two boxes in the upper left corner and lower right corner, respectively. The locations of the stations belonging to the nodes in the cluster with (b) the low attention scores and (c) the high attention scores are shown in the original map.}
\label{fig:attention_score}
\end{figure*}

%%% Graph Attention Network (GAT)의 어텐션 스코어 분석
In the graph attention network, each cell that contains several stations is deemed a (coarse-grained) node, and the attention score $\alpha_{ij}$ in Eq.~(\ref{eq:attention}) stands for the link weight between nodes $i$ and $j$, which stores the long-range correlation between spatially distant nodes. The attention score matrix, or the weighted adjacency matrix, is shown in Fig.~\ref{fig:attention_score}(a). The higher (lower) score $\alpha_{ij}$ represented as darker (lighter) red implies stronger (weaker) relatedness between two cells. We probe the two representative clusters of the lowest and highest weights encompassed by the purple and blue squares on the matrix, and mark the original positions of the stations in the map [Figs.~\ref{fig:attention_score}(b) and~\ref{fig:attention_score}(c)]. The stations belonging to the cluster with the lowest similarity are those that are almost distributed around the center. Otherwise, highly similar stations are scattered around the outskirts. It signifies that the outskirt regions do not have enough information to update node features by themselves. In other words, these regions need to get information from the other nodes (regions). That is why the outskirt regions have higher attention scores among them than the center regions of the city. Therefore, the network structure of our trained graph attention relates to the amount of information used between nodes.

\subsection{Prediction performance}
\label{subsec:performance}

% Fig: 지역 수요와 대여소 수요에 대한 예측 성능
\begin{figure}
\centering
\includegraphics[width=0.8\columnwidth]{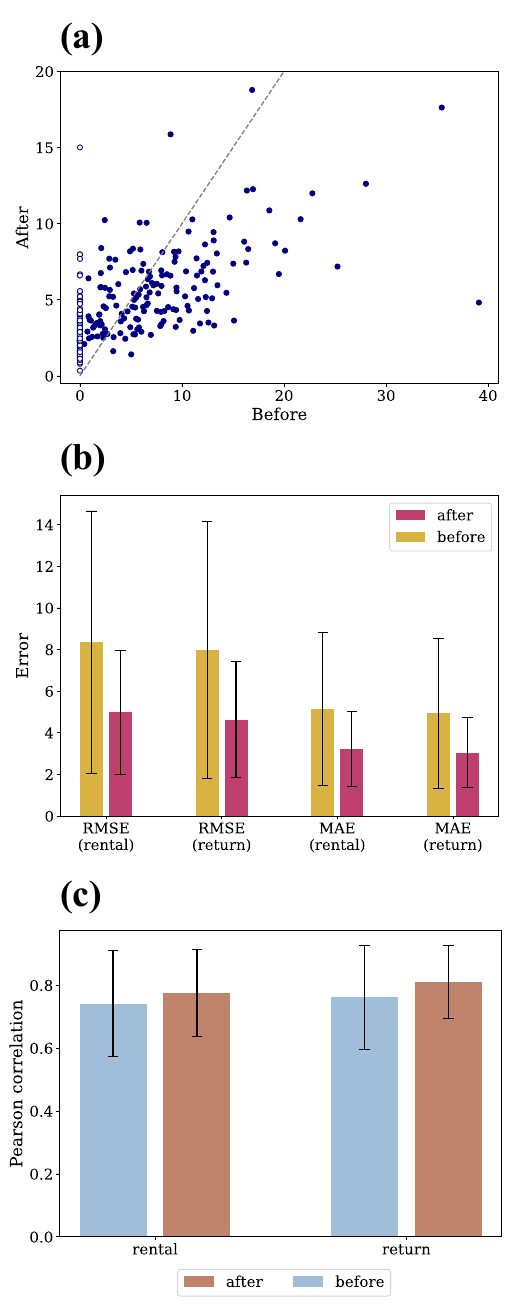}
\caption{The prediction result before and after using the cartogram– for the year 2019. (a) Scatter plot of $e^{\rm RMSE}_{i}$ for the rental for all cells, before and after applying the cartogram. The straight line stands for $y=x$. (b) The average errors $e^{{\rm RMSE}}$ and $e^{{\rm MAE}}$ for the rental and return.  (c) The average Pearson correlation between empirical $X_i(t)$ and prediction $\hat{X}_i(t)$ for every cell $i$. The height of the bar corresponds to the average value, and the error bar is a standard deviation.}
\label{fig:performance}
\end{figure}

%%% 지역 수요 예측
We evaluate the prediction performance of the ST-CGA model for the year 2019 by the following errors; the root-mean-square error $e^{\rm RMSE}_{i}$ and the mean absolute error $e^{\rm MAE}_{i}$ for a cell $i$ computed as
\begin{subequations}
    \begin{align}
        e^{\rm RMSE}_{i} &= \sqrt{\frac{1}{T}\sum_{t\in \mathbb{T}} [X_i(t) -\hat{X}_i(t)]^2}, \label{eq:rmse} \\
        e^{\rm MAE}_{i} &= \frac{1}{T}\sum_{t\in\mathbb{T}}|X_{i}(t)-\hat{X}_{i}(t)| \label{eq:mae},
        \end{align}
    \label{eq:error}
\end{subequations}
with $T=|\mathbb{T}|$. To confirm the effect of the cartogram in Sec.~\ref{subsec:cartogram}, we compare the errors in Eq.~(\ref{eq:error}) before and after applying the cartogram. As a representative case, we exhibit the scatter plot of $e^{\rm RMSE}_{i}$ for the rental in Fig.~\ref{fig:performance}(a), and one sees entirely the lower errors after applying the cartogram. The average errors in Fig.~\ref{fig:performance}(a) are shown as the leftmost bars in Fig.~\ref{fig:performance}(b), and the averages for the other cases (rental/return and error types) are illustrated as well. Using the cartogram for both rental and return cases results in more enhanced prediction than before applying the cartogram, supported by the about 1.6 times lower average errors: for a year, $e_{{\rm RMSE}}$ ($e_{{\rm MAE}}$) estimates the deviation from the empirical data with about eight (five) bicycles not using the cartogram and with about five (three) bicycles using cartogram. The cause for higher performance is expected to be the augmentation of the input data $X_i$ under the statistical quality control. We also measure the temporal Pearson correlation between empirical $X_i(t)$ and predicted $\hat{X}_i(t)$, and the correlation with the slightly higher average and reduced deviation tells us that our method grasps the temporal trend well. Figures~\ref{fig:performance}(b) and~\ref{fig:performance}(c) allow us to conclude that the use of the cartogram improves the prediction regardless of the type of demand (rental or return).

\subsection{Prediction for a new station}
\label{subsec:new_station}
% Fig: 신설 대여소 수요 예측 
\begin{figure}
\centering
\includegraphics[width=\columnwidth]{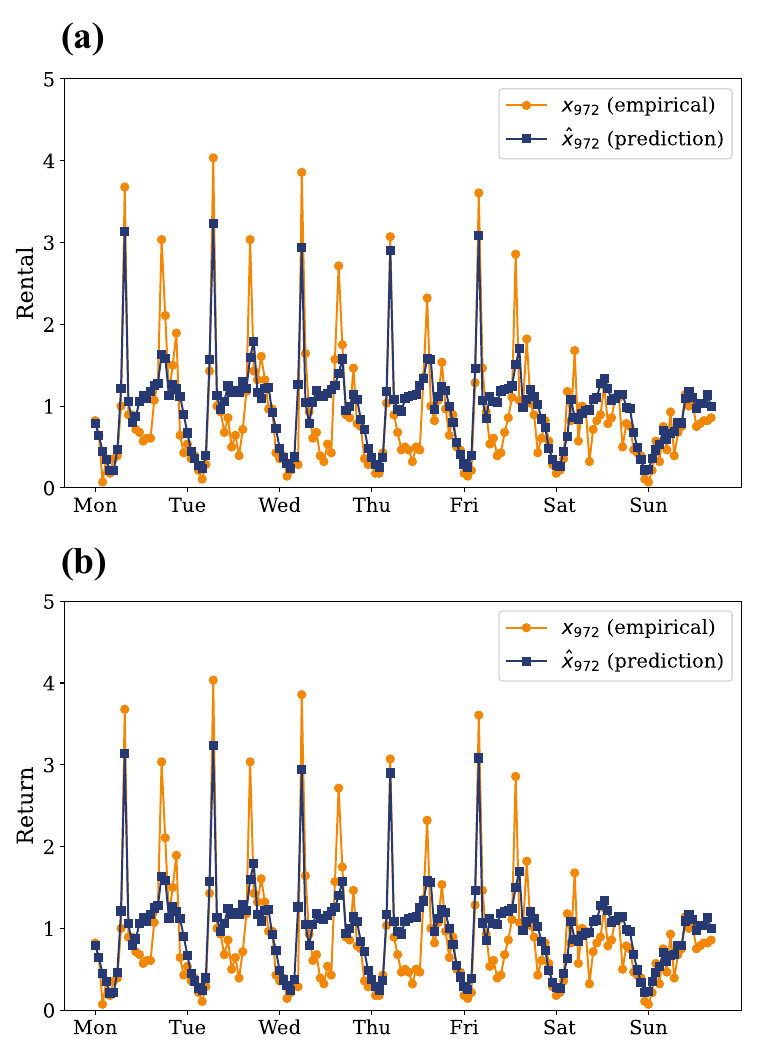}
\caption{The demand prediction of the newly installed station, $j=972$, for (a) the rental and (b) the return in 2019. The orange circle and blue square indicate the average values of the empirical data $x_{972}$ and the prediction $\hat{x}_{972}$.}
\label{fig:prediction_for_new_station}
\end{figure}

We deal with coarse-grained cells rather than individual stations due to the computation cost, so our ST-CGA model only provides the aggregated prediction of demand at the cell level. To guess the demand $\hat{x}_j(t)$ of an individual station $j$, we distribute the predicted demand of a cell $i$ to which the station belongs by the number of stations in the cell, under the assumption of the homogeneity of the demand within a cell supported by Fig.~\ref{fig:CV_P}. That is, 
\begin{equation}
    \hat{x}_j(t)\equiv \frac{1}{|\mathbb{R}_i|}\hat{X}_i(t) \,\,\,  \mathrm{ if }\   j\in \mathbb{R}_i.
\label{eq:xhat}
\end{equation}
We can obtain all $\hat{X}_i(t)$'s for all cells due to the cartogram without any exception, which allows us to guess the $\hat{x}_j(t)$ for a brand-new station $i$ that was not trained before. It opens up the possibility of the prediction for a brand-new station and enables the long-period, yearly prediction. The previous studies~\cite{Zhang2017regional_demand_example1, Kim2019regional_demand_example4, Li2019regional_demand_example2} exclude the brand-new demand caused by the newly installed stations, which results in the prediction in the limited period (a few weeks or months) to avoid the appearance of the new stations. At this point, we would like to emphasize that we surmount both of the short-period prediction and the exclusion of the new stations, by introducing the cartogram.

%%% 결과분석
The estimation results for a new station, labelled as $j=972$, are shown in Fig. \ref{fig:prediction_for_new_station}, for a week representatively. This station was newly installed and solely located in the cell marked by a red arrow in Fig.~\ref{fig:mean_x}(b) in 2019, so the cell in 2018 had no station and is shown as being empty in Fig.~\ref{fig:mean_x}(a). The prediction $\hat{x}_{972}(t)$ does not perfectly match the empirical value $x_{972}(t)$ but their maximal difference is less than one bicycle. Moreover, the quasiperiodic behavior shown in the empirical data is well reproduced in the prediction. The extent of predicting new demand that was not trained can be an indicator to assess the model's capability, beyond the prediction error~\cite{sutton2020identifying}.

%%% 기존 방법의 한계 (신설대여소 예측 불가능)와 카토그램을 활용한 예측
%The deep learning model cannot accurately predict demand for periods where demand is consistently zero during the training period. In our data, some regions have no demand during the training period but show demand during the test period. In such cases, the deep learning model struggles to predict demand for the test period because the model observed only zero demand during training. However, the cartogram method makes predicting demand for the new stations possible. The stations are uniformly distributed on the map using the cartogram method, and we observe that stations with similar characteristics are grouped together compared to before using the cartogram method (see Fig. \ref{fig:region std}). The uniform distribution of stations ensures that all regions have demand during both the training and test periods. Thus, the model does not solely observe zero demand during training and can learn the demand for new stations by incorporating information from nearby stations.

\section{Conclusion}
\label{sec:conclusion}

We have studied how to enhance the prediction of temporal patterns of rentals and return (demand) for public bicycles, as a representative open system. We have adopted the spatial-temporal convolution graph attention (ST-CGA) network as a prediction model, considering long-range interaction between distant stations. To facilitate new demand in brand-new stations and enhance the prediction performance, we have proposed and introduced the cartogram by iterating the Voronoi tessellation and modified the ST-CGA model by exploiting the batch attention and the update method of the node features in the graph attention. Due to the cartogram, we have achieved not only the original goal (prediction of new demand) but also long-period prediction (Fig.~\ref{fig:prediction_for_new_station}), and both have not been achieved in previous studies. The logic behind using a cartogram is to infer the unknown data by the average trend of the adjacent data in a cell, similar to the mean-field approach. In addition, our attempt to introduce batch attention instead of self-attention to the ST-CGA network has demonstrated improved performance. Using batch attention means that the temporal correlation as well as the spatial correlation also plays a crucial role in accurate prediction.

The fully connected (weighted) network considered herein may sometimes contain redundant information, which can impede higher accuracy or result in higher computational costs. Extracting sparse network structures requires appropriate filtering methods. Global thresholding is one basic method, but it comes with the challenge of selecting the threshold value, which could be addressed by threshold-free methods in machine learning~\cite{ziletti2018insightful}. When constructing sparse networks from the outset, understanding important regional features and direct relationships among regions is necessary, which is often difficult to achieve. In such cases, domain-knowledge-free preprocessing can be applied to construct the initial network~\cite{jha2018elemnet}. The reliability of such networks can be verified using graph neural networks or node2vec~\cite{ribeiro2023deep, lopes2022machine}.

The simple coarse-graining method shares the same spirit as the one with the cartogram, in the sense of inferring an unknown value by the average value of neighboring ones. As a reason why using the cartogram outperforms simple coarse-graining despite the similar sense, we conjecture that our cartogram ensures augmentation of training data points, i.e., the number of cells, with the qualitative level maintained. As seen in Fig.~\ref{fig:CV_P}, the input coarse-grained data $X_i$ after spreading out the stations significantly increases, i.e., by 100 data points. Furthermore, the adjacent stations not only gradually spread more uniformly but also focus on the spatial nearness of the stations possessing similar demand patterns. The deliberate rearrangement method results in the significant augmentation of the trainable data points (cells) with the statistics of correlation and dispersion within a cell maintained (Fig.~\ref{fig:CV_P}), although the variation of the number of stations can be enough to affect the statistics. Not only our suggested iteration method but also other spreading methods to keep or even strengthen the correlation and dispersion could work better on predictability. In addition, our framework does not require any system-specific features but spatio-temporal input data. Hence, while we adopt the demand of public bicycles as a representative of open systems, our framework with such cartogram considered may be applied to other open systems in general, such as $e$-scooter, taxi allocation, and the incident cases of disease spreading with migration effect~\cite{armbruster2023covid}.

\section{Acknowledgments}
\label{sec:acknowledgements}
The authors acknowledge Beom Jun Kim for the fruitful discussion. This research was supported by the National Research Foundation (NRF) of Korea through the Grant Numbers. NRF-2023R1A2C1007523 (S.-W.S.), NRF-2021R1C1C1007918 (M.J.L.). This work was also partly supported by Institute of Information \& communications Technology Planning \& Evaluation (IITP) grant funded by the Korea government (MSIT) [No.RS-2022-00155885, Artificial Intelligence Convergence Innovation Human Resources Development (Hanyang University ERICA)] (S.-W.S.). We also acknowledge the hospitality at APCTP.

\appendix
\section{The Self-Attention}
\label{seca:selfattention}

For a given temporal resolution $r$, we compose the input data set $\mathbb{X}^{r}(t)$ in Eq.~(\ref{eq:input_x}) based on the $\tau_r$-dimensional vector $\mathbf{X}_i^{r}(t;\tau_r)$ in Eq.~(\ref{eq:x_subset}). In the self- or batch attention, it is a first step to embed the $\tau_r$-dimensional vector in a hyperdimension $L$, so we firstly obtain a matrix $\mathbf{X}_{\rm emb}^{r}\in \mathbb{R}^{MN\times B\times L}$ in the embedding space ($\mathbb{R}$ being a set of real number). The embedding dimension $L$ is a hyperparameter, and we heuristically choose $L=32$. The $u$th row of $\mathbf{X}_{\rm emb}^{r}$ contains $d$ features of node $i$ at pivot time $t_b$ ($u=i+MNb$) but it is impossible to interpret the physical meaning of every feature in machine learnings.

Using the matrix $\mathbf{X}_{\rm emb}^{r}$, we achieve the node-feature matrix $\mathbf{y}$ by
\begin{equation}
    \mathbf{y}^{r} = \mathrm{softmax}\left[\frac{\mathbf{Q}^{r}\left(\mathbf{K}^{r}\right)^{T}}{\sqrt{L}}\right]\mathbf{V}^{r},
    \label{Aeq:outputy}
\end{equation}
where a query $\mathbf{Q}^r$, a key $\mathbf{K}^r$, and a value $\mathbf{V}^r$ matrices are computed as
\begin{align}
\mathbf{Q}^r &= \mathbf{X}_{\rm emb}^{r}\mathbf{W}_{\mathbf{Q}}^r, \nonumber\\
\mathbf{K}^r &= \mathbf{X}_{\rm emb}^{r}\mathbf{W}_{\mathbf{K}}^r, \nonumber \\
\mathbf{V}^r &= \mathbf{X}_{\rm emb}^{r}\mathbf{W}_{\mathbf{V}}^r,
\label{Aeq:qkv}
\end{align}
with importance matrices $\mathbf{W}$'s. The softmax function is a smoothing function defined as $\mathrm{softmax}(z_i)=e^{z_i}/\sum_j e^{z_j}$. Every entity of all importance matrices including those in the main text is initially assigned from a Gaussian distribution. The query, key, and value originally play distinct roles in the attention\cite{Vaswani2017selfattention}. Unlike the general attention mechanism, there is no distinction among their roles in the self- or batch attention, but we still use the terminologies for convention. The three matrices in Eq.~(\ref{Aeq:qkv}) have the same shape as $MNB\times L$, leading to the node feature matrix or the correlation matrix between every node at every pivot time, $\mathbf{y}^{r}\in\mathbb{R}^{MNB\times L}$. The $\sqrt{L}$ in Eq.~(\ref{Aeq:outputy}) is introduced to suppress the blow-up of the matrix product.

\end{document}